\documentclass[journal]{IEEEtran}

\usepackage{epsfig,epstopdf}
\usepackage{cite}
\usepackage{graphicx}
\usepackage[cmex10]{amsmath}

\usepackage{array}
\usepackage{amsfonts}
\graphicspath{ {./figs/} }
\usepackage{amssymb}

\usepackage[linesnumbered,ruled]{algorithm2e}

\usepackage{etoolbox}

\usepackage{threeparttable}
\usepackage{booktabs} 
\usepackage{multirow}
\makeatletter
\patchcmd{\@makecaption}
{\scshape}
{}
{}
{}
\makeatletter
\patchcmd{\@makecaption}
{\\}
{.\ }
{}
{}
\makeatother
\def\tablename{Table}

\usepackage{color, colortbl}
\definecolor{LightCyan}{rgb}{0.89,1,1}
\definecolor{DarkCyan}{rgb}{0.5,1,1}

\begin{document}
\title{Towards Efficient Processing and Learning with Spikes: New Approaches for Multi-Spike Learning}

\author{
%	Qiang~Yu,~\IEEEmembership{Member,~IEEE,} % <-this % stops a space
	Qiang~Yu,
	Shenglan~Li,
	Huajin~Tang,
	Longbiao~Wang,
	Jianwu~Dang,
	Kay~Chen~Tan,~\IEEEmembership{Fellow,~IEEE}

\thanks{This work was supported in part by the National Natural Science Foundation of China under Grant 61806139 and Grant 61876162, in part by the Natural Science Foundation of Tianjin under Grant 18JCYBJC41700 and Grant 19ZXZNGX00030, in part by the Peiyang Scholar Program of Tianjin University under Grant 2020XRG-0050, in part by the Shenzhen Scientific Research and Development Funding Program under Grant JCYJ20180307123637294, in part by the Research Grants Council of the Hong Kong SAR under Grant CityU11202418 and Grant CityU11209219. \textit{(Corresponding author: Qiang Yu.)}}% <-this % stops a space

\thanks{Q.~Yu, S.~Li, L.~Wang and J.~Dang are with Tianjin Key Laboratory of Cognitive Computing and Application, College of Intelligence and Computing, Tianjin University, Tianjin, China (e-mail: yuqiang@tju.edu.cn).}% <-this % stops a space

\thanks{H.~Tang is with the College of Computer Science and Technology, Zhejiang University, China.}

\thanks{J.~Dang is also with Japan Advenced Institute of Science and Technology, Japan, and Huiyan Technology (Tianjin) Co., Ltd., Tianjin, China.}

\thanks{K.C.~Tan is with the Department of Computer Science, City University of Hong Kong, and the City University of Hong Kong Shenzhen Research Institute.}

}


\markboth{}%
{Shell \MakeLowercase{\textit{et al.}}: Bare Demo of IEEEtran.cls for Journals}



\maketitle

\begin{abstract}
%\boldmath
Spikes are the currency in central nervous systems for information transmission and processing. They are also believed to play an essential role in low-power consumption of the biological systems, whose efficiency attracts increasing attentions to the field of neuromorphic computing. However, efficient processing and learning of discrete spikes still remains as a challenging problem. In this paper, we make our contributions towards this direction. A simplified spiking neuron model is firstly introduced with effects of both synaptic input and firing output on membrane potential being modeled with an impulse function. An event-driven scheme is then presented to further improve the processing efficiency. Based on the neuron model, we propose two new multi-spike learning rules which demonstrate better performance over other baselines on various tasks including association, classification, feature detection. In addition to efficiency, our learning rules demonstrate a high robustness against strong noise of different types. They can also be generalized to different spike coding schemes for the classification task, and notably single neuron is capable of solving multi-category classifications with our learning rules. In the feature detection task, we re-examine the ability of unsupervised STDP with its limitations being presented, and find a new phenomenon of losing selectivity. In contrast, our proposed learning rules can reliably solve the task over a wide range of conditions without specific constraints being applied. Moreover, our rules can not only detect features but also discriminate them. The improved performance of our methods would contribute to neuromorphic computing as a preferable choice.

%With the presented rules, neurons can successfully discover embedded features in a background sensory activity. The simplicity and efficiency of our algorithms would be beneficial for both software and hardware implementations.

\end{abstract}

% \fi
\IEEEpeerreviewmaketitle

\section{Introduction}

\IEEEPARstart{H}{uman} brain has shown remarkable abilities on various cognitive tasks such as recognition, decision making, learning and memory, while operates with an extraordinarily low consumption of power and a fast speed of cognition \cite{kandel2000principles,dayan2001theoretical,merolla2014million,thorpe1996speed}. The excellence of the brain has inspired increasing efforts being devoted to understanding the principles how it operates as well as applying those principles to endow artificial systems with a similar ability on information processing as the brain. 

The perceptron \cite{rosenblatt1958perceptron} is one of the earliest brain-inspired attempt to build artificial neurons to learn for recognition. Starting from the perceptron model, artificial neural networks (ANNs) have drawn a great amount of attentions in the trend of artificial intelligence (AI). Driven by advances in a class of techniques called deep learning, ANNs have been thriving with a great success in tackling problems across diverse fields including image and speech recognition, natural language processing, autonomous driving and bioinformatics \cite{lecun2015deep}. Despite of their popularity, one of the major criticisms for current deep learning methods comes from the lack of biological plausibility. Additionally, deep ANNs are almost always trained on very fast and power-hungry modern day supercomputers with Graphic Processing Units (GPUs), leading to a challenge of running these networks on low-power devices. Substantial efforts are invested to improve the efficiency of ANNs \cite{hubara2016binarized}. However, there is still a huge gap in the efficiency as is compared to their biological counterparts, let alone the superior cognitive abilities of the brain. Therefore, it is desired to develop networks which are efficient on one hand and biologically plausible on the other hand to a certain extent.

Neurons in ANNs and their biological counterparts differ at least in the way how they communicate with each other. Artificial neurons use analog values while biological ones take advantage of spikes. It is believed that discrete spikes play an essential role in efficient processing {\cite{kandel2000principles,thakur2018large,victor1997metric}. Inspired by neuroscience, preliminary neuromorphic approaches from both software and hardware have been introduced to harness the advantages of biological systems \cite{merolla2014million,fuller2019parallel,xia2019memristive,goodman2009brian,Liu2019STDP,zheng2017online,GewaltigNEST}. Efficiency is one of the key focuses especially considering the fundamental inefficiency and non-scalability of the von Neumann architecture \cite{merolla2014million}. However, the learning capabilities of current neuromorphic hardware are relatively simple and limited due to the complex dynamics of the spiking agents\cite{young2019review,thakur2018large,balaji2019mapping}. On the other hand, the complexity of the spike processing and learning in software restricts their implementations on hardware. Thus, the gap between the two worlds motivates our study in this work towards efficient processing and learning with spikes, while bearing in mind the simplicity for implementation and the feasibility for potential developments in practical applications.
%some degrees of biological plausibility for understanding operating principles of the brain.

The fundamental currency in nervous systems is spikes, where information could be carried by the number of spikes, their occurrence time or their shapes \cite{kandel2000principles,Panzeri10,rowan2016synapse}. In order to emulate abilities of biological neurons on processing spikes, spiking neuron models are developed, and are believed to possibly lead a new generation of ANNs \cite{gerstner2002spiking,maass1997networks}. Popular spiking neuron models, such as Hodgkin-Huxley model \cite{hodgkin1952quantitative,kang2016dynamic}, Izhikevich model \cite{izhikevich2003simple}, leaky-integrate-and-fire (LIF) model \cite{burkitt2006review} and spike response model (SRM) \cite{gerstner2002spiking}, are introduced with certain levels of resembling behaviors of biological neurons. These models differ from the degree on describing details of neuronal dynamics, and thus their complexities vary from one to another. Although LIF and SRM models are relatively less biologically plausible as compared to the others, their simpler forms and ease of processing make them nearly the most commonly used spiking neuron models for brain-inspired or neuromorphic computing \cite{yu2013rapid,memmesheimer2014,gutig2016}.

Putting the processing units aside, how spikes can be used for information transmission still remains unclear, which restricts developments of spiking neural networks (SNNs) for a broad range of applications \cite{yu2016spiking}. The most two popular coding assumptions are the rate and the temporal codes \cite{kandel2000principles,dayan2001theoretical,YuNCS,Panzeri10,gutig2014spike}. A spike train conveys information with its number of spikes (or firing rate) under a rate code, while individual spike timing matters for representing information under a temporal one.  
The rate code is simple and robust to inter-spike-interval noise as it ignores the temporal structure of the spike train \cite{stein2005neuronal,london2010sensitivity}.
Such a rate code enhances the similarity between the non-spiking artificial neurons in ANNs and the spiking ones in SNNs, and thus rendering comparable performance in recognition tasks \cite{zheng2017online,cao2015spiking}. 
On the other hand, the temporal code has a high information-carrying capacity as a result of making full use of the temporal structure \cite{hopfield1995pattern,borst1999information}.
This makes the temporal code an appealing one for efficient processing \cite{mostafa2017supervised,yu2013rapid}. Most spiking frameworks or learning systems solely rely on a single coding scheme, but cannot be generalized, limiting their capabilities of utilizing advantages of different codes, as well as of exploring processing principles of the biological systems.

Unavoidably, the learning capability is an essential characteristic that is required for building cognitive artificial neural systems. The learning determines how neurons adapt their synaptic efficacies in response to the inputs in a way such that they could fit the environment to solve certain cognitive tasks. Therefore, we mainly focus on the learning in this work due to this importance.

Inspired by neuroscience, various learning rules have been discovered and developed in recent years. Hebbian learning is one of the earliest principles describing how neuronal connections are modified \cite{song2000competitive}, and it can be simply stated as ``neurons that fire together, wire together." Increasing experimental observations demonstrate that synaptic modification depends on tight temporal correlations between the spikes of pre- and post-synaptic neurons, leading to a temporally asymmetric form of Hebbian learning, the spike-timing-dependent plasticity (STDP) \cite{song2000competitive,bi2001synaptic,dan2004spike}. STDP enables neuron to process information in an unsupervised way \cite{masquelier2007unsupervised,masquelier2008spike}, but its dependence on temporal contiguity could limit its applicability \cite{gutig06,YuNCS,zheng2017online}.

Different supervised learning rules have been developed to train spiking neurons. The tempotron is an efficient learning rule that trains neurons to make decisions by binary behavior of firing or not, being reminiscent of the perceptron but with an additional time dimension being involved \cite{gutig06}. The binary response of the tempotron could constrain neuron's ability to fully utilize the temporal structure of the output \cite{yu2018spike}, steering efforts to a family of learning rules which can train neurons to fire at desired times \cite{bohte02spikeprop,ponulak2010supervised,florian2012chronotron,mohemmed2012span,YuQ2013PSD,memmesheimer2014}. However, how to construct an instructor signal with precise timings is challenging for both artificial and biological systems \cite{yu2016spiking,hu2018efficient}. Moreover, most of these supervised spike learning rules are designed for a temporal code, limiting their generalization to other schemes such as a rate one \cite{yu2018spike,brette2015philosophy}.

Recently, a supervised multi-spike tempotron (MST) rule \cite{gutig2016} is developed to train a neuron to fire a desired number of spikes, which empowers it to discover sensory features embedded in a complex background activity. This kind of multi-spike learning rule provides a new way for processing information under a broad range of coding schemes and has shown good performance on some sound recognition tasks \cite{yu2019robust}.
Improved modifications have been developed in \cite{yu2018spike,yu2018iconip}, along with detailed evaluations of different properties as well as theoretical proofs on convergence and robustness.
However, their complexity with respect to both learning and processing would limit their applicability to a large-scale and efficient neuromorphic developments.
In this work, we will continue to contribute towards this supervised multi-spike learning with simplicity, efficiency and capabilities of information processing bearing in mind. We focus more on extending the potential applicability of multi-spike learning rules by providing efficient alternatives.
The significance of our major contributions can be highlighted in the following aspects.

\begin{itemize}
	\item 
	A simplified LIF neuron model is introduced for efficient processing of spikes, making it valuable for both software and hardware implementations. This is significantly important especially with considerations of the highly complex nonlinear dynamics of a spiking model.
	Additionally, an event-driven scheme, where computation is driven by spikes, is described in our framework, further benefiting the efficiency for both processing and learning of spikes. 

	\item 
	We propose two new approaches for multi-spike learning, namely efficient multi-spike learning (EML) and an alternative relying only on neuron's current response (named as EMLC where `C' stands for `current'). The efficient performance of our learning rules together with their simplicity and computational capabilities contribute to build large scale neuromorphic systems which could potentially drive a paradigm shift on processing towards more brain-like.

	\item 
	We evaluate the performance of our learning rules on a broad range of typical tasks including efficient processing, multi-category classification and robustness, with comparisons to other baseline methods, demonstrating the advanced performance of our work. Our results thus can further provide useful reference for applied developments of neuromorphic systems.

	\item 
	The ability of spike learning rules on feature detection from background activities is evaluated with a specific interest due to its importance on perception. Notably, we re-examine the unsupervised learning with STDP on this task, observing a new finding about loss of detection after sufficient learning. On the contrary, our proposed rules show better performance on not only detection but also discrimination in a more challenging task. These preferable performances make our algorithms a potential tool for processing temporal information as well as for better understanding computational principles of the brain.
	
\end{itemize}

The remainder of this paper is structured as follows. Section~\ref{sec:Methods} introduces our proposed approaches for spike processing and learning. Section~\ref{sec:experiments} then shows our experimental results, followed by discussions in Section~\ref{sec:discuss}. Finally, we conclude our work in Section~\ref{sec:Conclusion}.

%Although an increasing number of experiments have been shown in various nervous systems \cite{london2010sensitivity,kandel2000principles,gollisch2008rapid,butts2007temporal} to support different codes, it is still arguable whether the rate or temporal code dominates information coding in the brain \cite{masuda2002bridging,gutig2014spike,yu2018spike}.


%\hfill January 11, 2007

\section{Methods}
\label{sec:Methods}
In this section, we will introduce the methods proposed in our work for spike processing and learning. Firstly, we describe an efficient neuron model, followed by descriptions of an event-driven scheme. Then, two new supervised multi-spike learning rules are proposed. Additionally, we also introduce the STDP rule that is used as a benchmark in our feature detection task.

\subsection{Neuron Model}

The simplicity of LIF and SRM neuron models makes them the most commonly used ones in neuromorphic computing. An LIF neuron model can be mapped to SRM with certain defined spike response functions \cite{gerstner2002spiking}. Therefore, we start with a current-based leaky integrate-and-fire neuron model due to its simplicity and analytical tractability. Following a typical description of an LIF \cite{gerstner2002spiking}, our neuron model is given as
\begin{equation}
\label{Eq:neuron_dv}
 \dfrac{dV}{dt} = -\frac{1}{\tau}V(t) + I_\mathrm{in}(t) + I_\mathrm{out}(t)
\end{equation}
where $\tau$ represents the time constant of neuron's membrane potential, $V$, $I_\mathrm{in}$ and $I_\mathrm{out}$ model the inputs from pre-synaptic neurons and firing reset dynamics, respectively. The units of all parameters except $\tau$ are set to 1 for a general description. We set the two inputs in a simple form as
\begin{equation}
\label{Eq:Iin}
I_\mathrm{in} (t) = \sum_{i=1}^N w_i\sum_{t_i^j\leq t} \delta(t-t_i^j)
\end{equation}
\begin{equation}
\label{Eq:Iout}
I_\mathrm{out} (t) = - \vartheta \sum_{t_\mathrm{s}^j<t} \delta(t-t_\mathrm{s}^j)
\end{equation}

Here, $\delta(t)$ is a unit impulse function where its value is 1 at $t=0$ and 0 elsewhere. It represents the occurrence of a spike from either pre-synaptic neurons ($t_i^j$) or neuron's own output ($t_\mathrm{s}^j$). $N$ and $w_i$ denote the number of pre-synaptic channels and their corresponding synaptic efficacy, respectively. $\vartheta$ represents the firing threshold of the neuron where a spike will be elicited whenever the neuron's membrane potential crosses it.

\begin{figure}[!htb]
	\centering\includegraphics[width=0.45\textwidth]{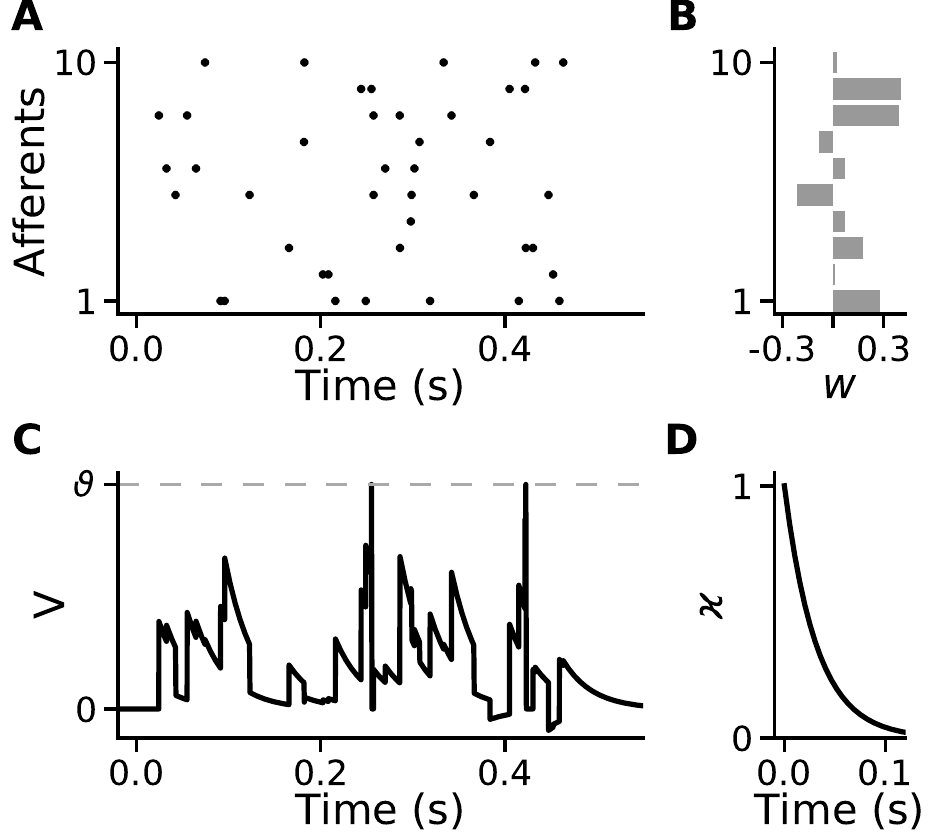}
	\caption{Demonstration of the evolving dynamics of the spiking neuron model. \textbf{A}, an exemplary spike pattern which contains 10 afferent neurons firing certain numbers of spikes (denoted by dots) across time. \textbf{B}, the corresponding synaptic weights ($w$) of a postsynaptic neuron receiving spikes from these afferents. \textbf{C}, the resulting membrane potential dynamics of the neuron in response to the pattern in \textbf{A}. The dashed line represents the firing threshold, $\vartheta$. \textbf{D}, demonstration of the post-synaptic potential kernel, $\varkappa$.}
	\label{Fig:neuron}
\end{figure}

Integrating Eq.~(\ref{Eq:neuron_dv}) with substitutions of $I_\mathrm{in}$ and $I_\mathrm{out}$, we get a form of SRM as
\begin{equation}
\label{Eq:neuron_srm}
V(t) = \sum_{i=1}^N w_i\sum_{t_i^j\leq t} \varkappa(t-t_i^j) - \vartheta \sum_{t_\mathrm{s}^j<t} \varkappa(t-t_\mathrm{s}^j)
\end{equation}
where $\varkappa (t) = \exp(-t/\tau)$ is an exponential kernel describing the influence of spikes on membrane potential. $\varkappa (t)$ is causal and thus vanishes for $t<0$. Notably, kernel $\varkappa$ is resulted from our model description in Eq.~(\ref{Eq:neuron_dv}), but it can be generalized to other choices with the basic form of Eq.~(\ref{Eq:neuron_srm}) being preserved.

A neuron continuously integrates afferent spikes into its membrane potential,
and generates output spikes whenever a firing condition is matched.
As can be seen from Fig.~\ref{Fig:neuron}, each afferent spike will result in a post-synaptic potential (PSP), whose peak value is controlled by the synaptic efficacy, $w$. In the absence of input spikes, neuron's membrane potential will gradually decay to the rest level, $V_{rest}$, and we set it to 0 here. Whenever the membrane potential crosses neuron's firing threshold, an output spike is elicited, followed by a reset dynamics.

Our neuron model shares certain similarity with that in other studies \cite{gutig06,gutig2016,yu2018spike} where a double-exponential kernel $K(t)$ is adopted as
\begin{equation}
K(t) = V_0\left[\exp{\left(-t/\tau_\mathrm{m}\right)}-\exp{\left(-t/\tau_\mathrm{s}\right)}\right]  
\label{Eq:kernel_double}
\end{equation}

Here, $V_0$ is a constant parameter that normalizes the peak of $K(t)$ to unity. $\tau_\mathrm{m}$ and $\tau_\mathrm{s}$ are the time constants for membrane and synaptic currents which are set to 20 and 5 ms, respectively, as a typically common choice.

\begin{figure}[!htb]
	\centering\includegraphics[width=0.43\textwidth]{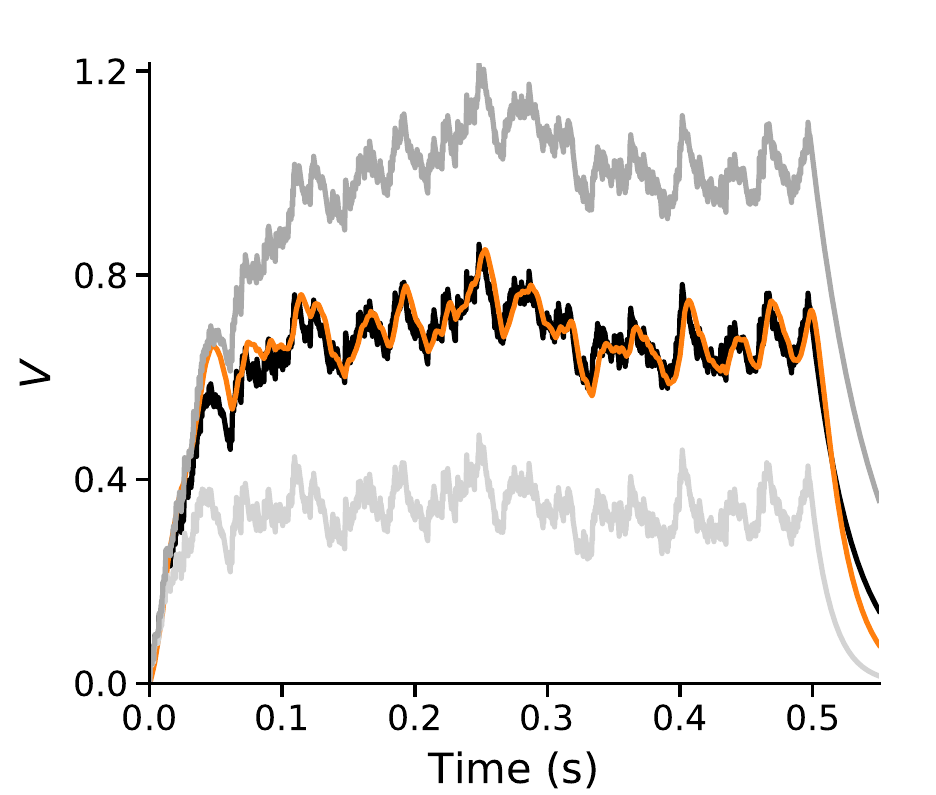}	
	\caption{Effects of neuron time constants on its membrane potential dynamics. All voltage traces are shown under a super-threshold scenario to relieve the firing nonlinearity. The orange line represents a typical neuron model equipped with a kernel $K(t)$ described in Eq.~(\ref{Eq:kernel_double}), where $\tau_\mathrm{m}$ and $\tau_\mathrm{s}$ are set to 20 and 5 ms, respectively. The other lines are traces of our neuron model with time constants of $0.5\tau$ (light gray), $\tau$ (black) and $1.5\tau$ (gray) where $\tau$ is set to result in a same temporal integration effect as $K(t)$ (see text for details).}
	\label{Fig:vtau}
\end{figure}

Importantly, the simplicity of our model can benefit both processing and learning of spikes, without loss of performance on cognitive tasks like recognition and feature detection. Spike response kernels reflect the ability of neurons to integrate information over time (see Fig.~\ref{Fig:vtau}). Larger $\tau$ covers longer duration of time, and thus results in collection of more information in the past. Due to this temporal integration over time, the effects of different kernels on neuron's membrane potential can be approximated following $\int K(t) dt = \int \varkappa(t) dt$, and thus we can set $\tau$ according to it.

\subsection{Event-Driven Scheme}

Following the approach in \cite{yu2018spike}, we adopt an event-driven computation for efficient processing. This event-driven approach is more efficient than a clock-based one because it does not depend on a step size for simulation, thus reducing computational operations to be linearly related to the total number of input spikes ($n$). Moreover, exact solutions can be obtained with the event-driven approach without constraints from the time resolution for simulation in a clock-based one. 

\begin{algorithm}
	\SetKwInOut{Input}{Input}
	\SetKwInOut{Output}{Output}
	
	\underline{function Response} $(S)$\;
	\Input{Spike pattern $S=\{t_k|k=1,2,...,n\}$}
	\Output{Number of output spikes $n_o$}

	initialization\;	

	\While{there is a new incoming spike $t_k$}{

	update membrane potential $V(t_k)$ according to Eq.~(\ref{Eq:neuron_event1})\;

	\While{$V(t_k)>\vartheta$}
	{
		update $V(t_k)$ with firing reset dynamics according to Eq.~(\ref{Eq:neuron_dv}) or Eq.~(\ref{Eq:neuron_srm})\;
%		$V(t_k) \leftarrow V(t_k)-\vartheta$\;
		$n_o \leftarrow n_o+1$\;		
	}

	}
	return $n_o$\;
	\caption{Event-driven computation}
\end{algorithm}

We consider a stream of input spikes $t_1\leq t_2 \ldots \leq t_n$ with $w_1,w_2,\ldots,w_n$ denoting
the corresponding synaptic weights associated with each spike.
According to Eq.~(\ref{Eq:neuron_srm}), neuron's membrane potential can be rewritten as
\begin{equation}
\label{Eq:neuron_event1}
V(t_k) = V(t_{k-1})\exp{\left( -\Delta_{k-1}/\tau\right)} + w_k
\end{equation}
where $\Delta_{k-1}=t_k - t_{k-1}$ denotes the inter-spike interval before the $k$-th input spike. Notably, Eq.~(\ref{Eq:neuron_event1}) describes the membrane potential without firing reset. If a neuron's potential is greater than its firing threshold, a reset dynamic will be involved to update its potential. Algorithm 1 shows the abstract scheme for our event-driven computation.

Notably, following a similar routine, the above event-driven approach can be easily transformed to a clock-based one where $\Delta_k$ is replaced by a fixed time step. In this work, we adopt the event-driven scheme.

\subsection{Multi-Spike Learning Rules}

Recently, a multi-spike tempotron (MST) rule is proposed to train neurons with a desired number of spikes \cite{gutig2016}, leading a new family of plasticity rules (here referred as multi-spike learning rules in this paper). Two threshold-driven plasticity (TDP1 and TDP2) rules are developed depending on the linear assumption around threshold crossing \cite{yu2018spike}, and improved performance has been demonstrated. Here, we continue to contribute to this new family of learning rules with efficiency mainly considered for processing and learning. In this work, we propose two new approaches, i.e. EML and EMLC, as follows.

\begin{figure}[!htb]
	\centering\includegraphics[width=0.42\textwidth]{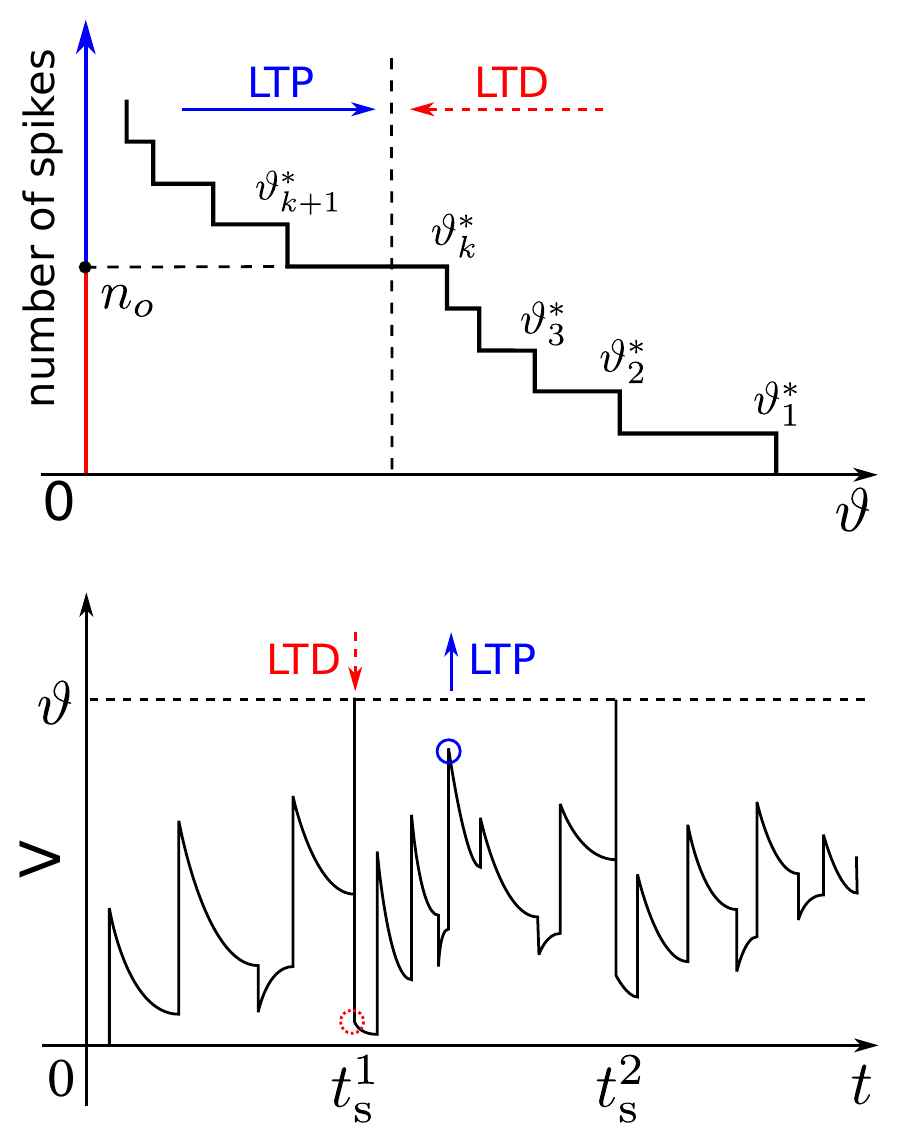}	
	\caption{Illustration of long-term potentiation (LTP) and long-term depression (LTD) plasticity rules of different multi-spike algorithms. Top, plasticity relies on critical thresholds, $\vartheta_k^*$, of the neuron's spike-threshold-surface (STS). Vertical dashed line denotes the threshold of the neuron. Bottom, plasticity only depends on the current dynamics of the neuron in response to a pattern. The red circle illustrates the minimum of reset potentials at all output spike times, $t_\mathrm{s}^j$. The blue circle denotes the maximum of sub-threshold potentials.}
	\label{Fig:learningrule}
\end{figure}

\subsubsection{The EML rule}
The first rule we proposed is called efficient multi-spike learning (EML) which is based on the spike-threshold-surface (STS), like MST \cite{gutig2016} and TDP \cite{yu2018spike}. STS characterizes the relation between neuron's actual output spike number and its firing threshold. A higher threshold value normally results in a lower number of output spikes. According to this property, critical thresholds $\vartheta_k^*$ can be highlighted as the position at which point neuron's output spike number jumps from $k-1$ to $k$. Therefore, modifications of these critical threshold values can result in a desired output spike number, but the challenge is how to change them.

Each critical threshold value $\vartheta_k^*$ corresponds to a voltage described by Eq.~(\ref{Eq:neuron_srm}), and thus it is a function of the synaptic weights $w_i$ and differentiable with respect to them. We define the maximum of subthreshold voltages for a given $\vartheta$ as $v_\mathrm{max}(\vartheta)$.
Consider a $\vartheta^*$ as the threshold, we assume there exists a $t^*$ such that $V(t^*)=v_\mathrm{max}(\vartheta^*)=\vartheta^*$. There could exist a number of output spikes that occur before $t^*$, and thus complicates the derivative evaluations \cite{gutig2016,yu2016spiking}. In our method, we find the dependence through previous output spikes can be neglected. This is because that for any preceding output spikes $j$, $\exists \xi>0$ such that $V(t^j_\mathrm{s})-\vartheta>\xi$. A sufficiently small change on $w$ will hardly affect $t^j_\mathrm{s}$. The derivative of $\vartheta^*$ with respect to $w_i$ is denoted as $\vartheta^{*'}_i$, and can thus be given as
\begin{equation}
\label{Eq:eml}
\vartheta^{*'}_i = \frac{\partial V(t^*)}{\partial w_i} 
= \sum_{t_i^j\leq t^*} \varkappa(t^*-t_i^j)
\end{equation}

According to Eq.~(\ref{Eq:eml}), a training method can thus be developed to adapt critical thresholds via changes of synaptic efficacies. Among many possible objectives, we choose one of the simplest that only considers $\vartheta^*_{n_o}$ and $\vartheta^*_{n_o+1}$ with $n_o$ denoting the actual output spike number. The supervised signal is the difference between the number of $n_o$ and $n_d$. The target is to train the neuron to elicit a desired number of spikes, $n_d$. The learning rule can be given as
\begin{equation}
\label{Eq:learning_eml}
\Delta w =
\begin{cases}
-\lambda \frac{d\vartheta^*_{n_o}}{d w}       & \quad \text{if } n_\mathrm{o}>n_\mathrm{d} \\
\lambda \frac{d\vartheta^*_{n_o+1}}{d w}  & \quad \text{if } n_\mathrm{o}<n_\mathrm{d}\\
\end{cases}
\end{equation}
where $d\vartheta^*_{k}/d w$ represents the directive evaluation calculated according to Eq.~(\ref{Eq:eml}), and $\lambda$ is a learning rate that controls the step size of each adaption.
The essential idea of this rule (see Fig.~\ref{Fig:learningrule}) is to decrease (increase) the critical values that are bigger (smaller) than $\vartheta$ with an LTD (LTP) process if a neuron fails to elicit a desired number of spikes.

\subsubsection{The EMLC rule}

The rules of MST, TDP and also as-proposed EML are all based on STS, and thus we refer them as STS-based rules. These methods depend on evaluations of critical thresholds as well as their derivatives with respect to synaptic efficacy, which is complex and could thus slow down the processing. Our preliminary attempt \cite{yu2018iconip} has demonstrated great improvement on efficiency by combining both the tempotron and PSD rules. Here, we further our study by proposing a new approach where only neuron's current states of response are considered. We name this rule as EMLC (here `C' stands for `current').

Intuitively, one quick way to change the output spike number is to directly modify the voltages that are close to the neuron's threshold. Following this idea, we choose the time point at the maximum subthreshold voltage to perform LTP, and denote this time point as $t_\mathrm{LTP}$. For the LTD process, we select the time point, $t_\mathrm{LTD}$, where the voltage after firing reset is the minimum among all output spikes. An illustration of the plasticity is shown in Fig.~\ref{Fig:learningrule}. The EMLC rule can thus be formalized as 
\begin{equation}
\label{Eq:learning_emlc}
\Delta w =
\begin{cases}
-\lambda \frac{\partial V(t_\mathrm{LTD})}{\partial w}       & \quad \text{if } n_\mathrm{o}>n_\mathrm{d} \\
\lambda \frac{\partial V(t_\mathrm{LTP})}{\partial w}  & \quad \text{if } n_\mathrm{o}<n_\mathrm{d}\\
\end{cases}
\end{equation}

According to Eq.~(\ref{Eq:learning_emlc}), a neuron can thus learn to fire a desired number of spikes based on its current states rather than STS.

\subsection{Unsupervised STDP Rule for Comparison}
Here, we introduce the unsupervised STDP learning rule adopted as a baseline in the subsequent task of feature detection. As the unsupervised STDP learning rule is widely studied in spike-based processing \cite{masquelier2008spike,masquelier2009competitive} and demonstrated capability of detecting features from background activities, so we choose it for a clear comparison with our supervised multi-spike method.

Different from approaches in \cite{masquelier2008spike,masquelier2009competitive} where `nearest spike' approximation is applied for the learning, we use a more general STDP learning rule where every pair of pre- and post-synaptic spikes will contribute a synaptic change (see Fig.~\ref{Fig:stdp}).

\begin{figure}[!htb]
	\centering\includegraphics[width=0.42\textwidth]{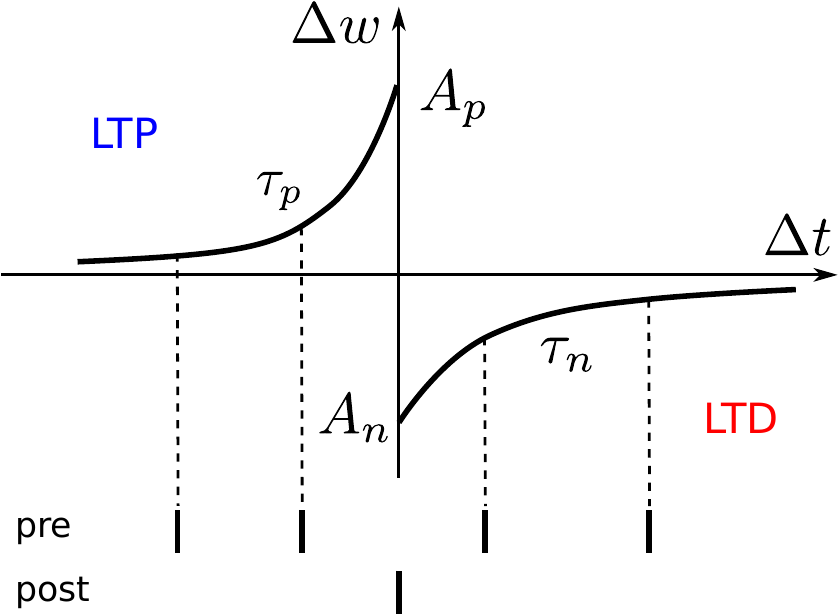}
	\caption{Demonstration of spike-timing-dependent plasticity (STDP). The weight modulation, $\Delta w$, depends on the time difference between the pre- and post-synaptic spike timings, $\Delta t=t_\mathrm{pre}-t_\mathrm{post}$. The top presents the learning window, while the bottom shows exemplary spikes from both pre- and post-synaptic neurons.}
	\label{Fig:stdp}
\end{figure}

The basic STDP rule can be formalized as
\begin{equation}
\label{Eq:learning_stdp}
\Delta w =
\begin{cases}
A_p \exp(\Delta t/\tau_p)       & \quad \text{if } \Delta t \leq 0 \\
-A_n \exp(-\Delta t/\tau_n)  & \quad \text{if } \Delta t > 0\\
\end{cases}
\end{equation}
where $\Delta t=t_\mathrm{pre}-t_\mathrm{post}$ denotes the time difference between a pair of pre- and post-synaptic spikes. $A_p$ and $A_n$ are the modulation magnitudes for LTP and LTD, respectively. $\tau_p$ and $\tau_n$ are the corresponding time constants of the STDP learning window.

\subsection{Momentum}
A momentum scheme could accelerate the learning \cite{gutig06}, and thus it is applied in our study.
The actual performed synaptic update $\Delta w$ is composed of two parts: the current modification, $\Delta w^\mathrm{current}$, that is determined by the corresponding learning rules, and a fraction of the previous applied update $\Delta w^\mathrm{previous}$. Therefore, in each error trial, the resulting synaptic update is as
\begin{equation}
\label{Eq:momentum}
\Delta w = \Delta w^\mathrm{current} + \mu \Delta w^\mathrm{previous}
\end{equation}
where $\mu \in [0, 1]$ is the momentum parameter determining the fraction of the previous update.

\section{Simulation Results}
\label{sec:experiments}

In order to better benchmark performances of our learning rules, we show simulation results in this section including derivative evaluation, learning efficiency and capabilities for classification and feature detection, etc. The default setups for the number of connected pre-synaptic afferents and the learning rate are as: $N=500$ and $\lambda=10^{-4}$, respectively. Neuron's synaptic efficacies (weights) are initialized with both the mean and standard deviation being set to 0.01. Input spike patterns are generated over a time window of $T=500$ ms with each afferent neuron firing at a Poisson rate of $r_\mathrm{in}=4$ Hz over $T$. Different setups from the default will be stated otherwise.
Our experiments were performed on a platform of Intel E5-2650@2.20GHz with two-processor Intel(R) Core CPU and 16GB main memory.

\subsection{Derivative Evaluation}

In addition to the rationale of our proposed EML rule introduced above, here we show its derivative evaluation with a similarity metric where the evaluation is compared to the theoretical derivative whose value is approximated numerically. Following the approach in \cite{yu2018spike}, the theoretical derivative is calculated as
\begin{equation}
\frac{\vartheta^*(w_i + \xi) - \vartheta^*(w_i)}{\xi}
\end{equation}
where $\xi$ is an infinitesimal change on weight. The similarity metric between two vectors, e.g., $\vec{x}$ and $\vec{y}$, is calculated as
\begin{equation}
\cos(\theta)=\frac{\vec{x} \cdot \vec{y}}{|\vec{x}| |\vec{y}|}
\end{equation}
Here, $\vec{x}$ and $\vec{y}$ are two vectors representing the derivative evaluation of a method and the theoretical one, respectively.

\begin{figure}[!htb]
	\centering\includegraphics[width=0.42\textwidth]{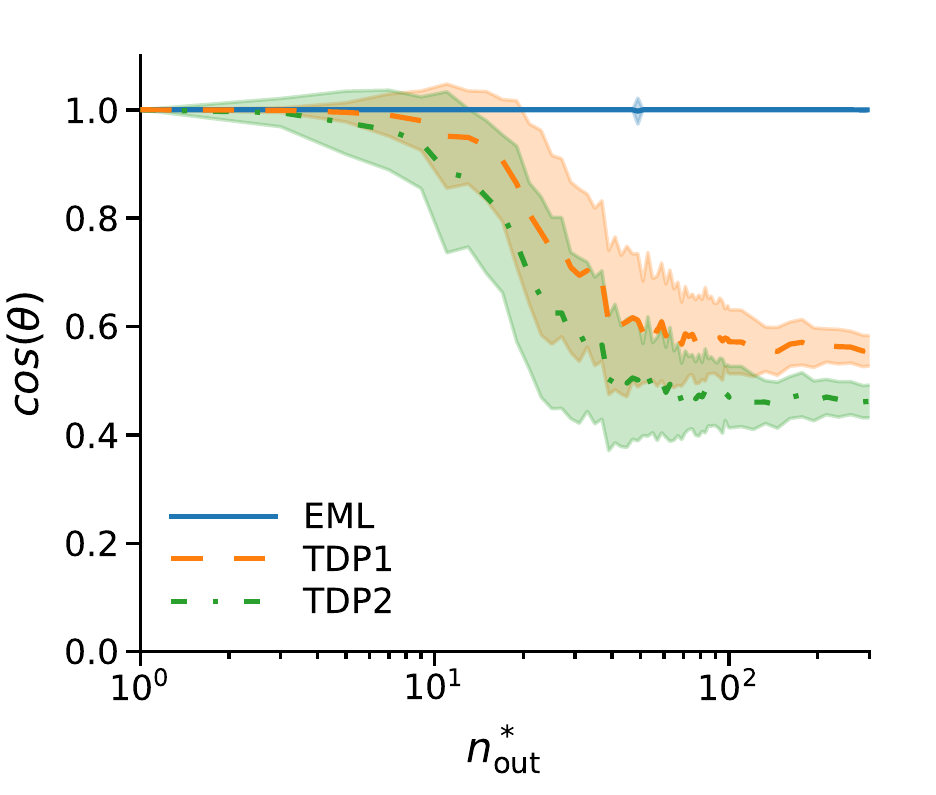}
	\caption{Derivative evaluations of different STS-based rules. Their similarities, $\cos(\theta)$, with the theoretical one are presented against different critical output numbers, $n_\mathrm{out}^*$. Each line and the corresponding shaded area denote the mean and standard deviation over 100 independent evaluations.}
	\label{Fig:derivative}
\end{figure}

In Fig.~\ref{Fig:derivative}, we present the evaluations of TDP1, TDP2 and EML rules with this metric. As can be seen from the figure, both TDP1 and TDP2 methods will slowly diverge from the theoretical evaluation when $n_\mathrm{out}^*$ increases. The vector angles will stay around $60^\circ$ when $n_\mathrm{out}^*$ is large, indicating a positive component contribution towards the same direction as the theoretical one. We will denote TDP1 as TDP in our following experiments to provide detailed benchmarks. Notably, our proposed EML rule provides a perfect match with the theoretical one due to the characteristics of our model. In our method, the impact of synaptic efficacy via preceding output spikes on $\vartheta^*$ can be neglected, resulting in a simpler and yet accurate evaluation of derivatives, and thus benefiting both the processing and learning.

\subsection{Learning Efficiency}

In this part, we will examine the learning efficiency of different multi-spike learning rules, including MST \cite{gutig2016}, TDP \cite{yu2018spike}, EML and EMLC.
In the task, these learning rules need to train neurons to elicit a desired number of output spikes, and their training efficiencies are recorded for comparison. 

\begin{figure}[!htb]
	\centering\includegraphics[width=0.45\textwidth]{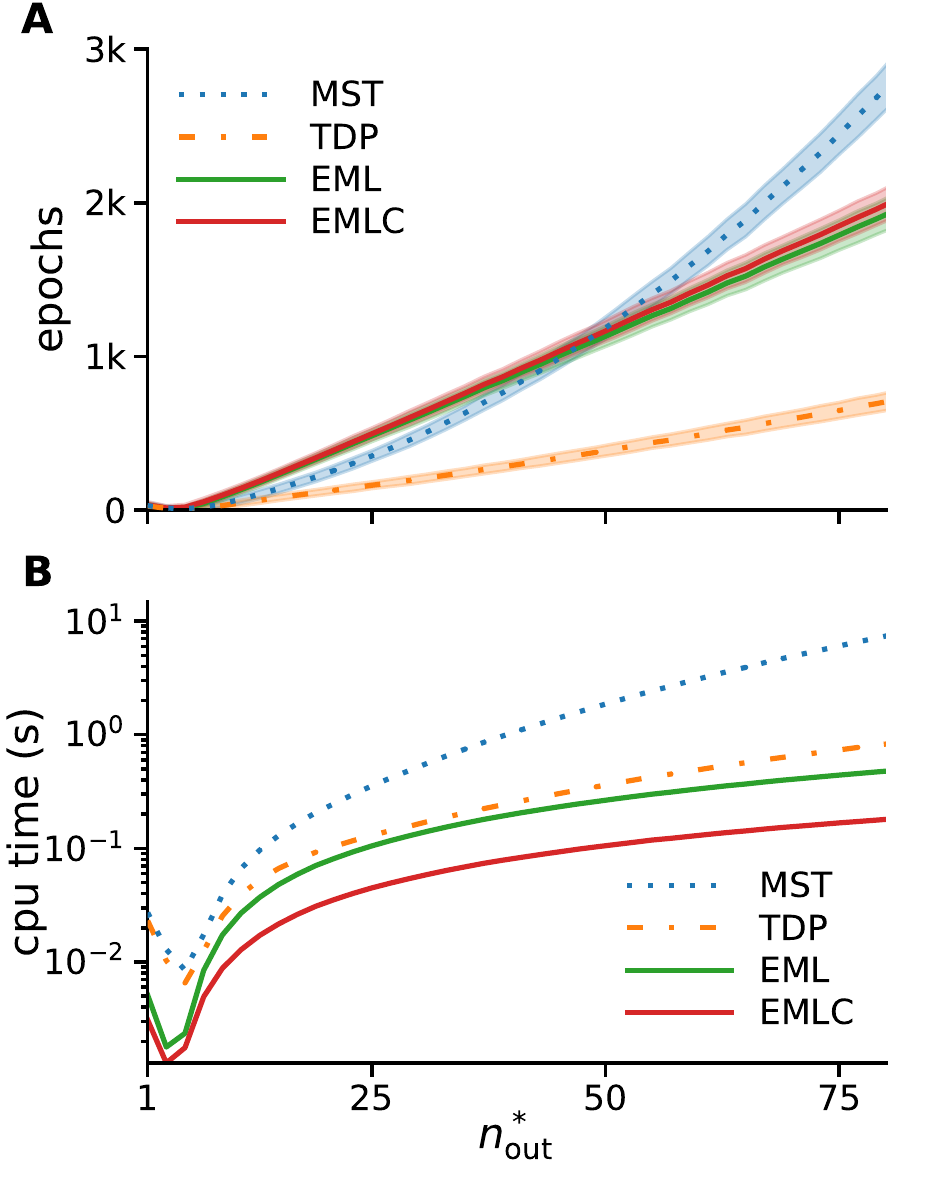}
	\caption{Efficiency of different multi-spike learning rules. \textbf{A} and \textbf{B} show the convergence epochs and the corresponding cpu running time, respectively. Neurons are trained with different rules to elicit certain output spike numbers, $n_\mathrm{out}^*$, in response to a spike pattern. Data were collected over 100 independent simulation runs.}
	\label{Fig:learnspeed}
\end{figure}

In the first experiment, we set a relatively high firing rate of $r_\mathrm{in}=6$ Hz to increase the computational load. A neuron was trained with an input spike pattern being presented to it one time after another until it fires a desired spike number, $n_\mathrm{out}^*$. Each pattern presence is denoted as one epoch. The total training epochs and cpu times are recorded until successful learning. Fig.~\ref{Fig:learnspeed} shows the learning efficiency versus different choices of $n_\mathrm{out}^*$. When $n_\mathrm{out}^*$ is small, the difference of these rules with respect to training epochs is small. When $n_\mathrm{out}^*$ increases, TDP uses the least number of epochs to finish the learning while MST is relatively slow as compared to others.
When we consider the actual execution time on cpu, both of our proposed rules, i.e. EML and EMLC, are faster than those baseline rules. Our EMLC is the most efficient one, with an average over $10\times$ faster than MST.

\begin{figure}[!htb]
	\centering\includegraphics[width=0.45\textwidth]{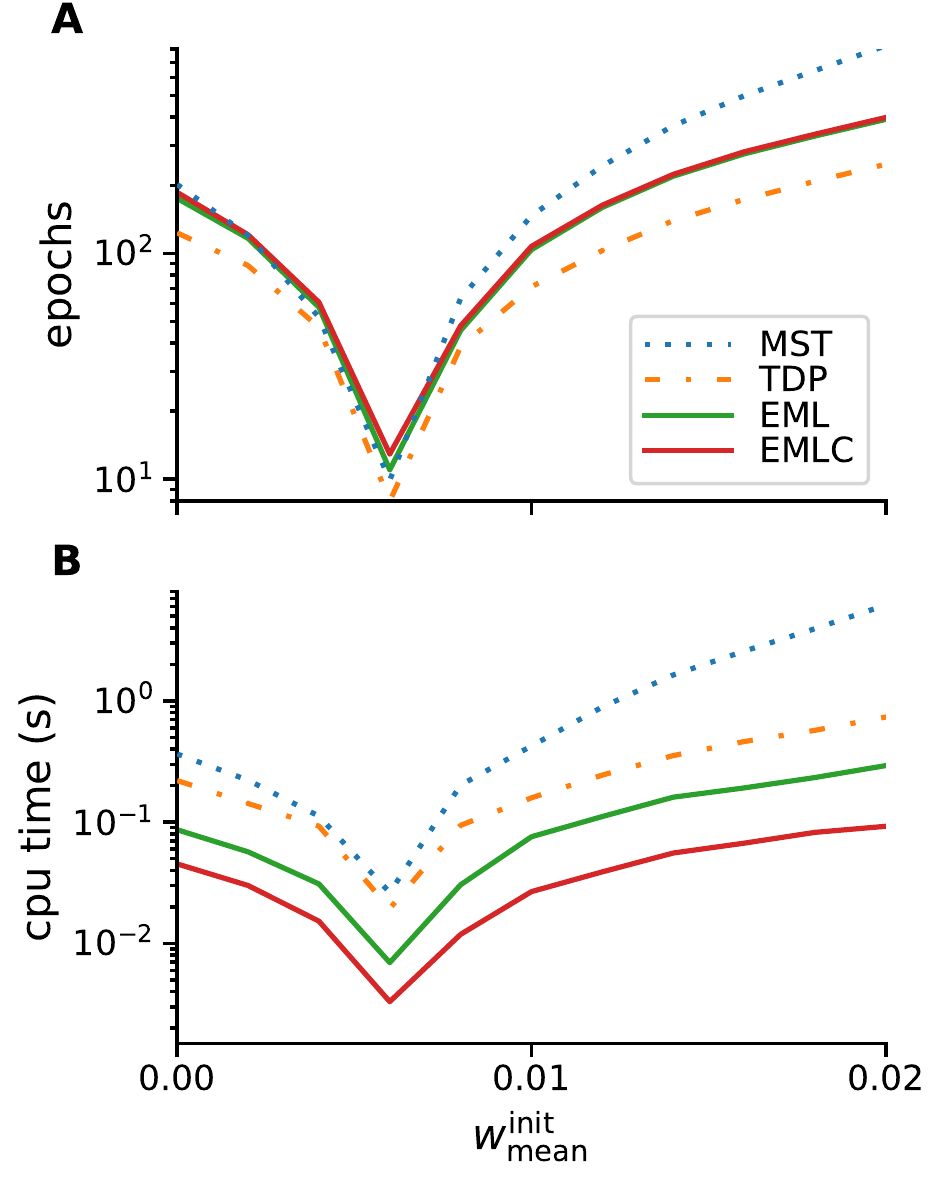}
	\caption{Learning efficiency against different initial setup conditions, $w_\mathrm{mean}^\mathrm{init}$. \textbf{A} and \textbf{B} show the convergence epochs and cpu execution time of different learning rules. Data were averaged over 100 runs.}
	\label{Fig:winit}
\end{figure}

Different initial setups will result in a different number of output spikes, and can thus affect the learning. To examine the learning reliability over different initial cases, we conduct our second experiment where different initial mean weights are used to initialize synapses. In this task, we set $r_\mathrm{in}=10$ Hz and $T=1.0$ s to further increase the computational load of each neuron. Every neuron is trained to fire 10 spikes. Similarly, training epochs and cpu times are recorded until successful learning.

As can be seen from Fig.~\ref{Fig:winit}, the learning speeds of all multi-spike rules change with different initial mean weights due to the incremental updating characteristics of the learning. Different mean weights will result in different initial output spike numbers, and thus the closer this value to the desired, the faster the learning. Similar to the findings in Fig.~\ref{Fig:learnspeed}, MST is relatively slow as compared to the others. Both of our proposed methods outperform the other two in terms of computational efficiency, i.e. cpu time. EMLC is the fastest one among these methods, and it is more than $10\times$ efficient as is compared to MST on average.

\subsection{Learning to Classify Spike Patterns}

Classification is a typical cognitive capability of most artificial intelligent agents \cite{sen2016binarization}.
In this experiment, we study the ability of different rules on discriminating spike patterns of different categories. Here we design a multi-category problem with 3 classes as an example. 
The neuron parameters are the same as previous except that the mean and standard deviation of initial weights are set as 0 and 0.001, receptively. Additionally, a momentum scheme \cite{gutig06,yu2018spike} with $\mu=0.9$ is applied to accelerate the learning. We perform two different tasks.

\begin{figure}[!htb]
	\centering\includegraphics[width=0.45\textwidth]{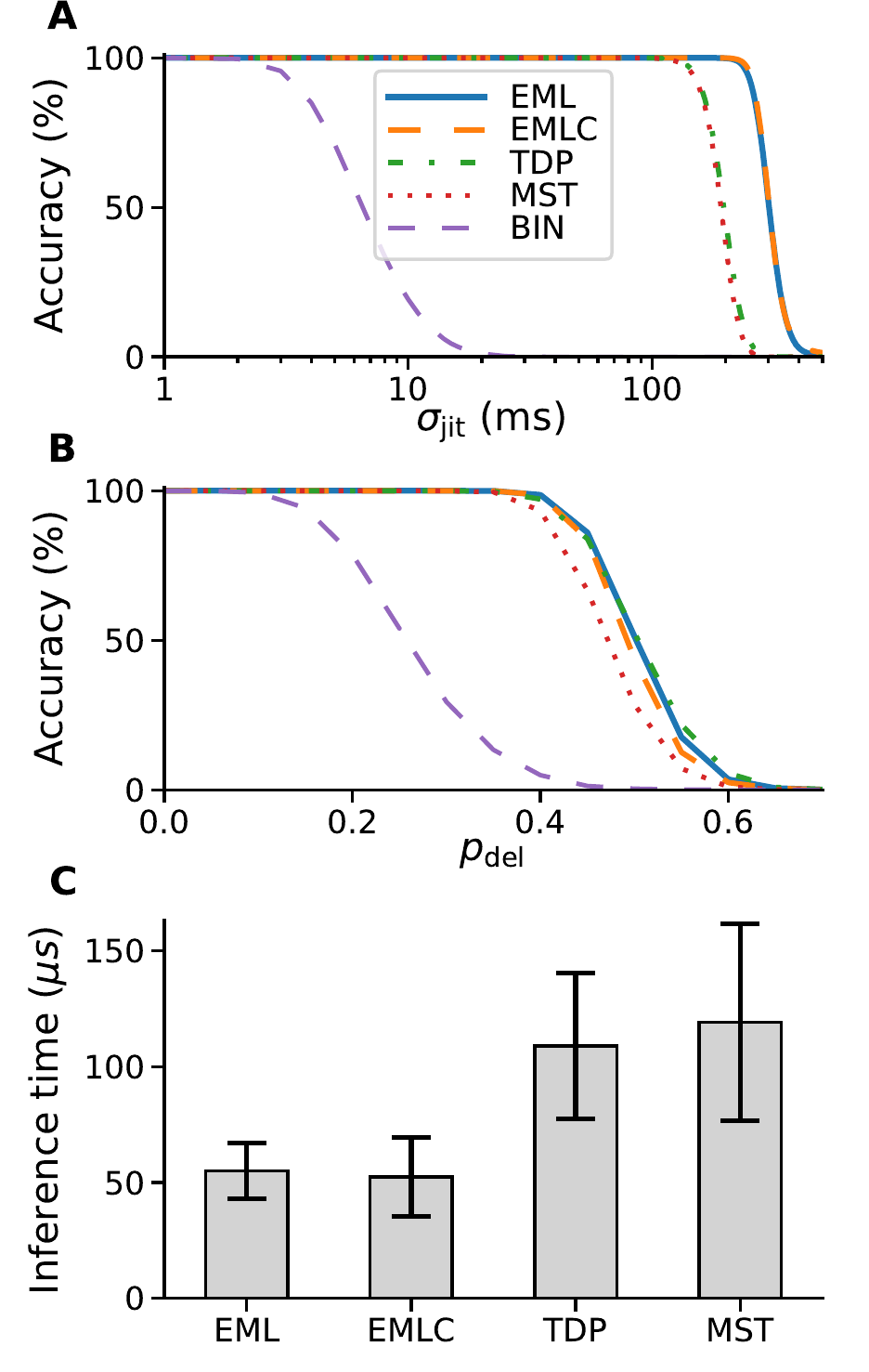}
	\caption{Learning performance of different rules on spatiotemporal spike pattern classification. \textbf{A} and \textbf{B} show the learning accuracy against spike jitter noise ($\sigma_\mathrm{jit}$) and spike deletion noise ($p_\mathrm{del}$), respectively. In addition to the multi-spike learning rules, the performance of the binary-spike tempotron rule (`BIN') is also presented. Neurons in \textbf{A} and \textbf{B} are trained with $\sigma_\mathrm{jit}=2$ ms and $p_\mathrm{del}=0.1$, respectively. \textbf{C}, inference time of neurons in \textbf{A} with both $\sigma_\mathrm{jit}=2$ ms and $p_\mathrm{del}=0.1$ being imposed. Data were collected over 100 independent simulations.}
	\label{Fig:spike_cls}
\end{figure}

In the first task, we consider a spatiotemporal spike pattern classification where each pattern is generated with $r_\mathrm{in}=2$ Hz resulting in an average of one spike per synaptic channel over the time window \cite{florian2012chronotron,mohemmed2012span,YuQ2013PSD}. 
Three templates are randomly generated and then fixed after that. Spike patterns of each category are instantiated by adding two types of noise to the corresponding template: spike jitter noise $\sigma_\mathrm{jit}$ and spike deletion noise $p_\mathrm{del}$.
We use $\sigma_\mathrm{jit}=2$ ms and $p_\mathrm{del}=0.1$ to train neurons for different noise types.
Three neurons are trained with the multi-spike rules to elicit more than 20 spikes for their corresponding target category and keep silent for the others. A strict readout scheme is adopted to highlight the superiority of the multi-spike learning over the binary one. During the evaluation phase, one can choose different readout schemes to inference the category of the input pattern. Here, we simply select a decision spike number to be half as the one used for training. That is, only if the corresponding neuron fires more than 10 spikes, we refer it as a correct action, otherwise as wrong. Note that, other readout schemes could also be applied to further improve the performance such as a competing one \cite{yu2016spiking}, but we use a simple scheme here to solely highlight the learning capabilities of our methods.

As can be seen from Fig.~\ref{Fig:spike_cls}, all the multi-spike learning rules outperform the binary tempotron rule in terms of robustness, indicating the advantages of exploiting output temporal structure with multiple spikes. These multi-spike learning rules can tolerate more than 100 ms jitter and 40\% random spike deletion with a high accuracy (100\%) being preserved. Notably, our proposed learning rules, i.e. EML and EMLC, are more robust than the other multi-spike learning ones. Moreover, our methods are more efficient (over $2\times$) than the others with respect to the inference time after learning. There is no significant difference on the inference time between EML and EMLC since learning is not involved. This processing efficiency could be beneficial for low-power devices.

\begin{figure}[!htb]
	\centering\includegraphics[width=0.42\textwidth]{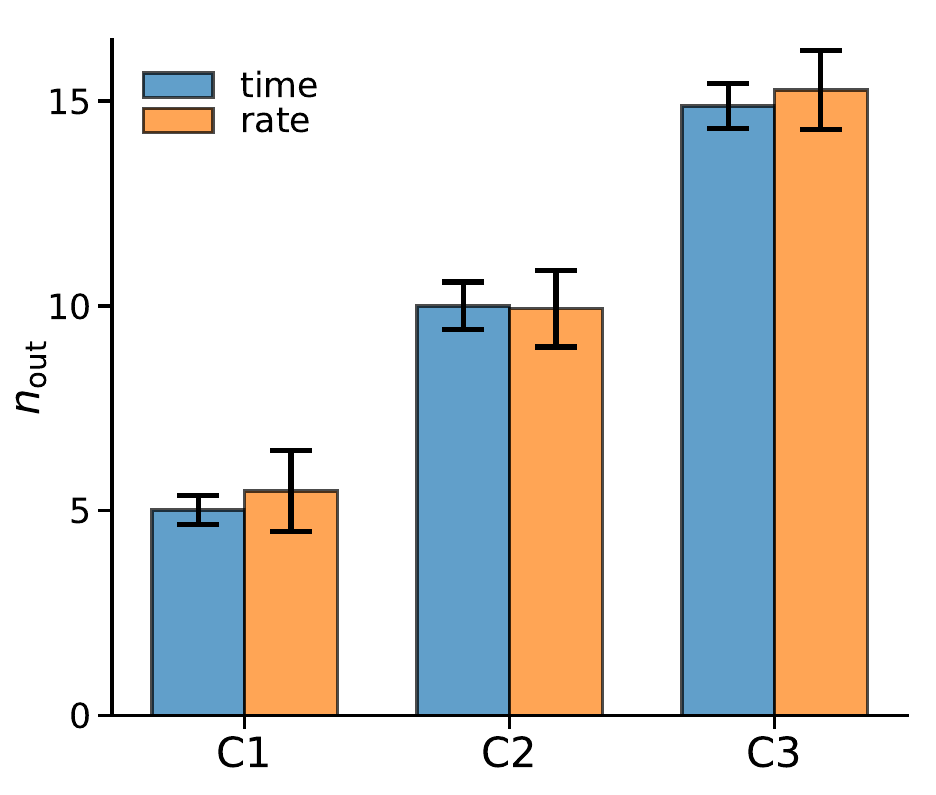}
	\caption{Capability of a single neuron to perform multi-category classifications for both time- and rate-based spike patterns. The neurons is trained to classify three categories by eliciting a number of spikes, $n_\mathrm{out}$, as: 5 (C1), 10 (C2) and 15 (C3). Data were collected over 1000 evaluations.}
	\label{Fig:multiclass}
\end{figure}

In the second task, we examine the ability of our EML rule to train a single neuron to solve the challenging multi-category classification. Following a similar experimental setup in \cite{yu2018spike}, we consider two different scenarios where a time- or rate-based coding scheme is used to generate spike patterns. The generation of time-based spike patterns is similar to the previous classification task. For the rate-based scenario, we randomly
generated 3 firing-rate templates where a random half afferents have a low firing rate of 2 Hz while the other half has 10 Hz. Each spike pattern is generated according to the Poisson process every time, and thus information is carried by the firing rates rather than precise spike timings. We train the neuron with desired numbers of spikes in response to different categories as: 5, 10 and 15.
Fig.~\ref{Fig:multiclass} shows that our learning rule can successfully train a single neuron to perform the multi-category tasks for both time- and rate-based spike patterns.
This highlights that our learning rule can be generalized to different coding schemes and might be applied to a broad range of situations where how spikes are used to code the information might be unclear.

\subsection{Feature Detection with STDP}

Previous studies show that neurons with STDP can detect features from background activities in an unsupervised way \cite{masquelier2007unsupervised,masquelier2008spike}, but the dependence of STDP on temporal contiguity could limit its capability. In order to provide a good benchmark for our methods on the feature detection task, we firstly re-examine the ability of the well-known STDP for unsupervised feature detection. The learning performance and its limitations are highlighted.

\begin{figure}[!htb]
	\centering\includegraphics[width=0.45\textwidth]{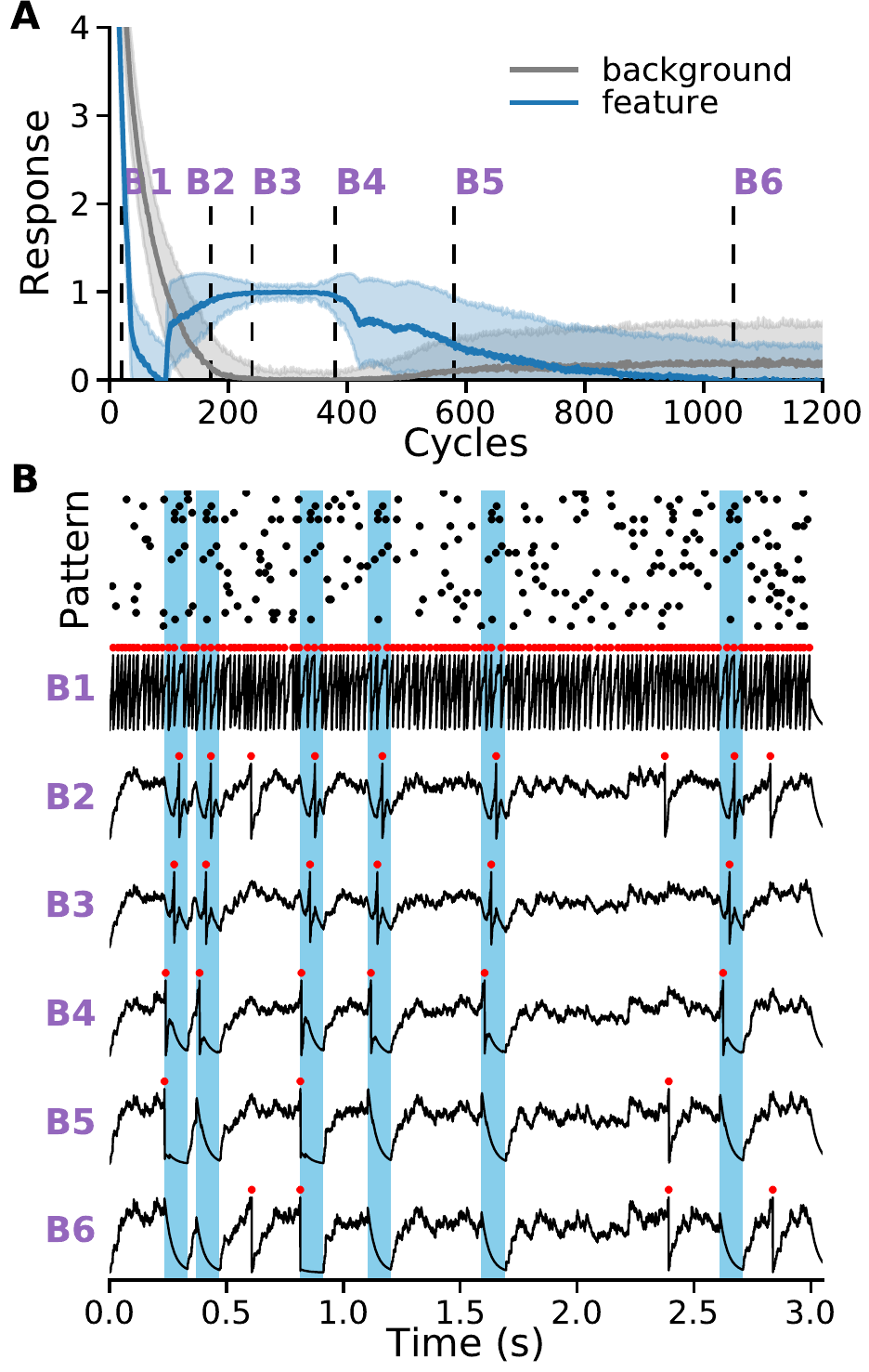}
	\caption{Feature detection with STDP. \textbf{A}, neuron's response to both background and feature spike patterns along the training cycles. Solid lines and shaded areas denote the mean and standard deviation, respectively. Data were collected over 1000 evaluations. Six typical phases (`B1'-`B6') were marked to demonstrate the behavior of the neuron along learning. \textbf{B}, the dynamics of the neuron, at different phases marked in \textbf{A}, in response to a sample spike pattern trial. Each black dot in the pattern represents a spike and only 4\% of the afferents are presented for a better visualization. Appearance of feature is highlighted with the shaded blue bars. `B1'-`B6' show the voltage traces and red dots represent the output spikes.}
	\label{Fig:lfstdp}
\end{figure}

We set $A_p=5\times 10^{-6}$, $A_n=0.72A_p$, $\tau_p=20$ ms, and $\tau_n=40$ ms, resulting in a domination of LTD over LTP as that in \cite{masquelier2008spike}. This domination is required for the success of STDP as it suppresses those non-selective spikes. Otherwise, the neuron will experience an explosion of spikes without it. We use a high initial weight setup with the mean and standard deviation being set to 0.05 and 0.01 respectively. This is because STDP depends on the appearance of output spikes and thus high initial values can result in more spikes to facilitate the learning.

We construct a random feature pattern with 4 Hz of afferent firing rate over a time window of 100 ms. This feature is randomly embedded by replacement in a background activity of 5 s (called a trial pattern here) with the same firing rate as the feature. We set the occurrence rate of the feature to 3 Hz. Then, a background firing noise of 1 Hz is added to finalize the trial pattern. Each training cycle consists of 10 trial patterns. 
In the evaluation phase, we use a specific scheme to reliably evaluate the neuron's response to spike patterns. Specifically, a background pattern ($P_\phi$) with a time window of 2 s is firstly generated, and then is fed to the neuron with a resulting output spike number ($R_\phi$) being recorded. A feature pattern (or background with the same length of time as the feature) is inserted in the middle of $P_\phi$, and then we record the neuron's response to it as $R_f$ (or $R_b$ accordingly). The difference between $R_f$ (or $R_b$) and $R_\phi$ is used as a response measurement. In this way, we can eliminate the reset effect on the neuron. Notably, we additionally constrain synaptic weights to fall between 0 and 0.1. This constraint is essential for the success of STDP, like that in \cite{masquelier2008spike,masquelier2009competitive}.

Fig.~\ref{Fig:lfstdp} shows the learning dynamics of STDP for unsupervised feature detection. Initially, the neuron fires at a high rate in response to both background and feature patterns (B1). Then, the neuron gradually depresses its response to background toward zero, and selectivity on feature pattern starts to appear (B2). The learning enters a plateau of selectivity (B2-B4). During this phase, neuron learns to find the starts of the feature pattern (B4), a characteristic of STDP which is similar to \cite{masquelier2008spike,masquelier2009competitive}. Differently, we find in our study that the neuron will gradually lose this selectivity for further learning (B4-B6). This is due to the domination of LTD which decreases synaptic efficacies to a level that neuron will rarely fire. Our new finding suggests that the domination of LTD is beneficial for the learning at the beginning on one hand, but probably a disaster for further learning on the other hand. Proper balance and tuning for the success of STDP are just required.

\subsection{Feature Detection with Multi-Spike Learning}

In this part, we examine the ability of our multi-spike learning rules for feature detection. We perform two different tasks. A momentum of $\mu=0.9$ is adopted to accelerate the learning \cite{gutig06,yu2018spike}.

\begin{figure}[!htb]
	\centering\includegraphics[width=0.45\textwidth]{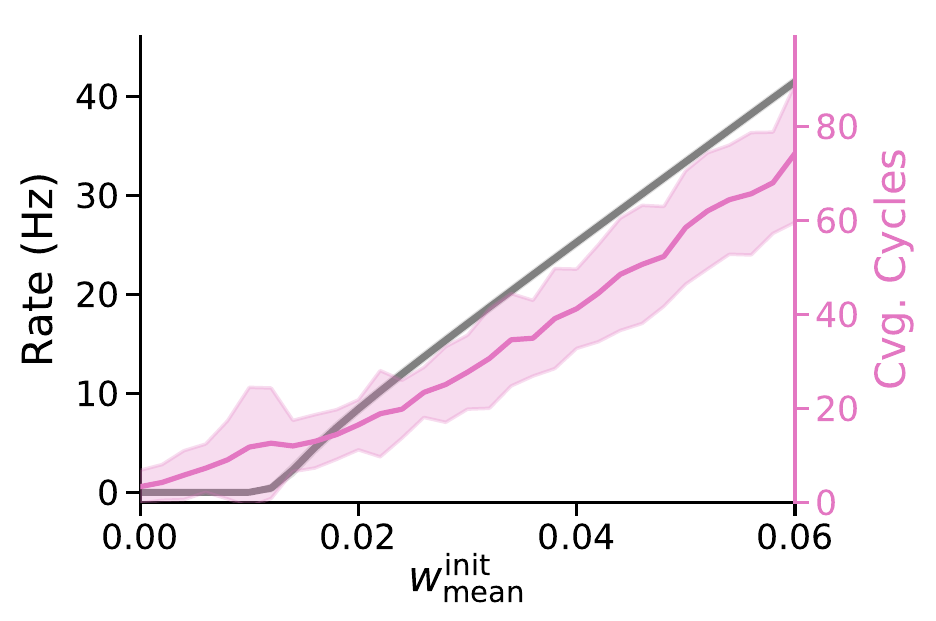}
	\caption{Feature detection with the EML rule under different initial conditions, $w_\mathrm{mean}^\mathrm{init}$. Both the initial firing rate of the neuron and the convergence cycles of the learning are presented. Data were collected over 100 independent runs.}
	\label{Fig:lfwmn}
\end{figure}

In the first task, our EML rule is applied to solve the same task as above with STDP. Differently, we do not impose any constraints on weights. Additionally, we evaluate our learning performance over a broad range of initial setups, $w_\mathrm{mean}^\mathrm{init}$ to examine the learning reliability. Fig.~\ref{Fig:lfwmn} shows the learning performance of our EML rule. Different values of $w_\mathrm{mean}^\mathrm{init}$ can result in a different initial firing rate ranging from 0 to tens of Hz. Our EML rule can successfully learn the task over all the given setups without any failure, suggesting the reliability of our learning rule. On contrast, this broad range of setups would be a disaster for STDP. For example, when there is no output spikes (with a small $w_\mathrm{mean}^\mathrm{init}$), there will be no learning for STDP at all. Differently, our learning rule can converge much faster than the unsupervised STDP learning.

\begin{figure}[!htb]
	\centering\includegraphics[width=0.43\textwidth]{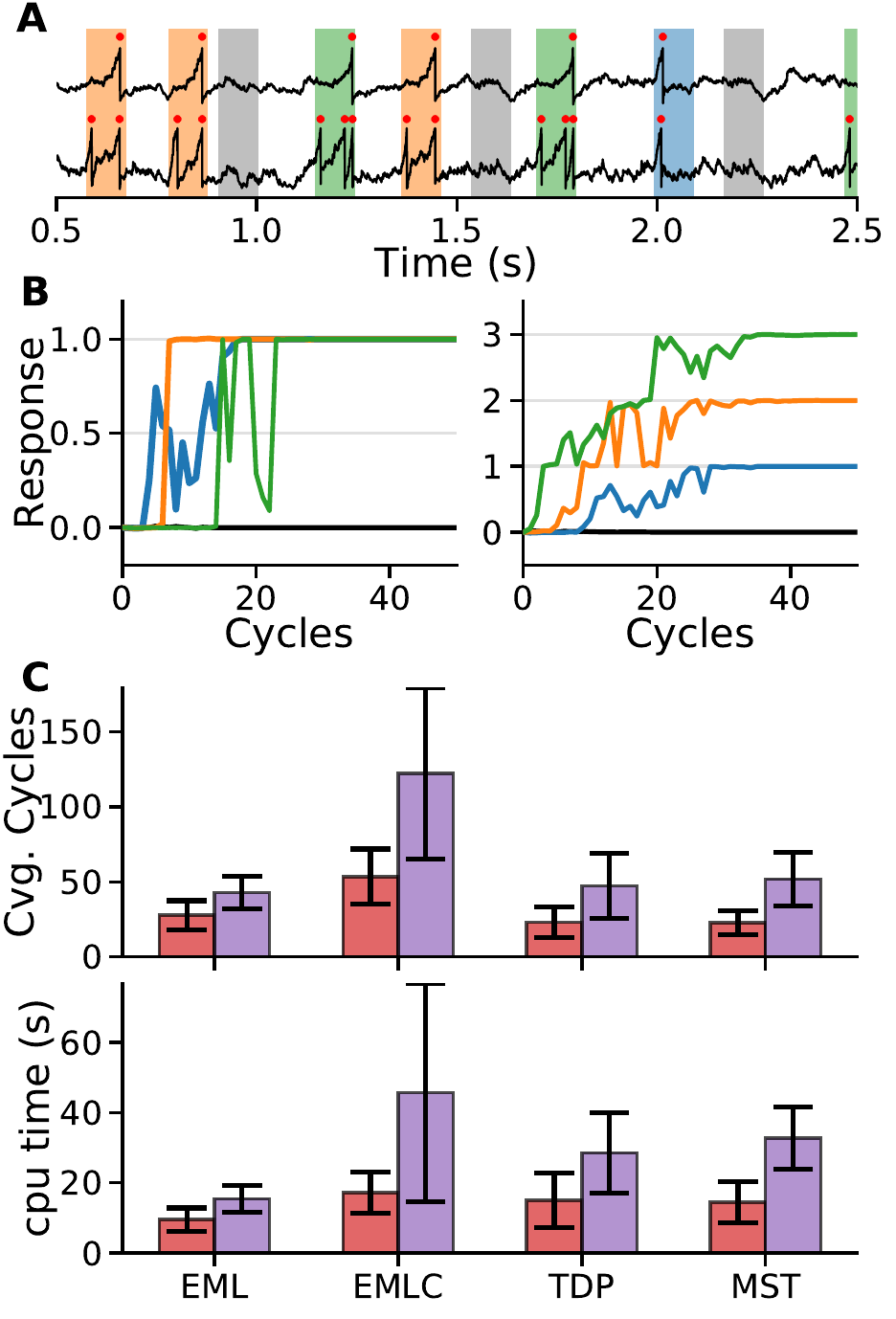}
	\caption{Multiple feature detection and recognition with multi-spike learning rules. \textbf{A}, neuron's membrane potential dynamics, after training (with the EML rule), in response to a spike pattern stream where both features (shaded blue, orange and green) and distractors (shaded gray) are embedded in a random background activity. The desired output spike number for each target feature is \{1, 1, 1\} (top) and \{1, 2, 3\} (bottom). \textbf{B}, the response of the neuron to both features (coded with the same color in \textbf{A}) and background (black) along learning. The left and right present the corresponding tasks in \textbf{A}. \textbf{C}, learning efficiency of different multi-spike rules for the tasks of \{1, 1, 1\} (red) and \{1, 2, 3\} (purple). The top panel shows the number of training cycles until convergence, and the bottom is the average cpu time of a training cycle. Data were collected over 100 independent runs.}
	\label{Fig:multif}
\end{figure}

In the second task, we consider a more challenging one where multiple activity patterns are embedded in the background. Similar to the task in \cite{yu2018spike}, we set 6 patterns with half being selected as features and the rest as distractors. Different from the previous trial generation, we first generate a background pattern of 2 s, and then randomly insert activity patterns into it. The occurrence number of each activity pattern over the trial is randomly generated by a Poisson process with mean of 3. The time window of the resulting trial pattern thus has a mean value of 3.8 s. A background noise of 1 Hz is then added to the trial pattern. Each training cycle contains 100 random trial patterns. We consider two scenarios where the neuron is trained to fire a desired number of spikes in response to each feature pattern as: \{1, 1, 1\} and \{1, 2, 3\}. The neuron is required to be silent to both distractors and background. 
The total desired output spike number $n^*_\mathrm{out}$ in response to a trial pattern is $n^*_\mathrm{out} = \sum_{i} c_i d_i$ where $c_i$ is the occurrence number of the $i$-th feature and $d_i$ is the corresponding desired output spike number for this feature pattern. We record the convergence cycle at the point where the difference between neuron's actual and desired output response to all activity patterns as well as background is less than 0.05 within 10 consecutive cycles.

Fig.~\ref{Fig:multif} shows the learning performance. The multi-spike learning can successfully detect multiple feature patterns that are embedded in a complex background where both noise and distractors are presented (Fig.~\ref{Fig:multif}A). Importantly, our learning rule successfully demonstrates the ability of discrimination in addition to detection within tens of training cycles (Fig.~\ref{Fig:multif}A,B).
Fig.~\ref{Fig:multif}C shows the learning efficiency of different rules on the feature detection task. The EMLC is relatively weak as compared to the others in terms of convergence cycles, but is comparative to that of TDP and MST in terms of cpu execution time. The challenging task of feature detection involves a long duration of noise and distractors where it is slightly harder to explore repetition with the neuron's current states (EMLC) than that with STS (the others). As a result, our EMLC takes longer cycles to converge.
Our EML is the most efficient one among all these multi-spike learning rules. It is around $2\times$ faster in terms of cpu time than that of TDP and MST for both scenarios. This efficiency makes our learning a potential candidate to benefit low-power neuromorphic computing.

\section{Discussions}
\label{sec:discuss}

Central nervous systems use spikes for both information transmission and processing \cite{kandel2000principles,dayan2001theoretical}. Spikes are believed to play an essential role in low-power consumption which would be of great importance to benefit devices such as mobiles and wearables where energy consumption is one of the major concerns \cite{merolla2014million,xia2019memristive}. The efficiency is now one of the major bottlenecks of deep learning methods \cite{lecun2015deep}, and thus attracts more attention to neuromorphic computing \cite{YuNCS,fuller2019parallel,zheng2017online,wade2010swat,woods2018fast}. In this work, we make our contributions towards this direction.

We first start with the spiking neuron model as it is the building block for processing spike information.
Different spiking neuron models have been proposed with different degrees for describing details of biological systems \cite{hodgkin1952quantitative,izhikevich2003simple,burkitt2006review,gerstner2002spiking}.
Complex models \cite{hodgkin1952quantitative,izhikevich2003simple} have a relatively high biological plausibility, but are computationally inefficient as a result. Simpler models \cite{burkitt2006review,gerstner2002spiking} are just more favorable in neuromorphic computing \cite{YuNCS,wade2010swat,zheng2018sparse} due to their simplicity while being capable of processing spikes.
In our work, we present a further simplified neuron model where unit impulse function is used to describe the effects of both synaptic input and firing output on neuron's membrane potential. We convert this model to an SRM form based on which an event-driven scheme is then introduced for processing. Remarkable efficiency for both processing and learning is just obtained as a result of our model (e.g. Fig.~\ref{Fig:learnspeed}B, \ref{Fig:winit}B and \ref{Fig:spike_cls}C). Note that, our neuron model and learning can also benefit a step-based computational scheme \cite{goodman2009brian,GewaltigNEST}, making it a favorable choice for efficient processing and learning. 

The new family of multi-spike learning rules \cite{gutig2016,yu2018spike} are recently developed to train neurons with a desired number of spikes rather than precise timings.
These learning rules could be preferable to others \cite{bohte02spikeprop,YuQ2013PSD,brader07} for making decision and exploring temporal features from the signals. This is because the multi-spike rules do not require precise timing to be specified as an instructor, and importantly can be generalized to different coding schemes \cite{yu2018spike}. We first develop an STS-based learning rule, i.e. EML, following a similar approach as in \cite{gutig2016,yu2018spike}. Importantly, our EML rule inherits the advantages of both MST and TDP, namely accuracy and efficiency of derivative evaluation, respectively (see Fig.~\ref{Fig:derivative}). Differently, the simplicity of our EML makes it better than the other two for processing and learning.
In order to further improve the processing efficiency, we propose the EMLC rule which has no dependence on STS, and thus accelerates the processing as a result. Our simulation results show that the EMLC is more efficient for association and recognition tasks (Fig.~\ref{Fig:learnspeed}, \ref{Fig:winit} and \ref{Fig:spike_cls}) as a result of STS avoidance, while the EML outperforms the EMLC in the feature detection task because critical thresholds could help to extract repeating information in a complex background.

Classification is a typical ability of a learning system that is widely studied in the field of neuromorphic computing \cite{wade2010swat,ponulak2010supervised,YuQ2013PSD,zheng2017online,brader07,zheng2018sparse,jeyasothy2018sefron}.
We examine the ability of our proposed rules on this task with multi-category being considered. Our learning rules are highly robust against a wide range of different noise types, and importantly they are more efficient than other learning rules (see Fig.~\ref{Fig:spike_cls}). Since a single neuron can be assigned to have different output spike numbers in response to different categories, our learning rules can thus empower single neurons to solve the multi-category tasks (Fig.~\ref{Fig:multiclass}). Importantly, this classification ability of our learning rules can be generalized to different spike coding schemes. Therefore, the efficiency, robustness and generalization of our learning rules make them a priority over others as a potential spike-based classifier. With a proper encoding scheme, our learning rules show a promising performance on some real-world classification tasks \cite{li2020new}.

Useful information is often embedded in the streams of sensory activities, making detection of feature information a challenging task (also called credit-assignment problem) \cite{gutig2016,schultz1997neural,friedrich2011spatio}. Numerous experiment data show that human brain leverages Bayesian principles to analyze sensory stimuli and infer the hidden states from the complex environment. Moreover, the unsupervised STDP rule has been theoretically proven as an approximation of  the Expectation-Maximization(EM) algorithm in machine learning,  which is often used for parameter estimation with hidden states such as Bayesian inference\cite{guo2019hierarchical,nessler2013bayesian}. Early studies demonstrate that STDP can find the starts of repeating feature patterns from a background in an unsupervised way \cite{masquelier2008spike,masquelier2009competitive}.
However, the success of STDP depends on certain experimental conditions such as domination of depression and high initial firing response. Additionally, our new results show that the detection would not be sustained if further learning occurs due to the domination of depression. Recent study also proves that unsupervised learning is fundamentally impossible for certain tasks \cite{locatello2018challenging}.
On the other hand, our supervised learning rules demonstrate a reliability in successfully detecting embedded features over a broad range of different setups (Fig.~\ref{Fig:lfwmn}). Importantly, the teacher signals are weak ones by only specifying the total number of spikes the neuron should elicit in response to a spike pattern. The neuron can then automatically detect the occurrence of each feature pattern. Moreover, discrimination of feature patterns can be accomplished in addition to detection with the multi-spike learning rules. Again, our proposed learning rule still outperforms the others in terms of efficiency (Fig.~\ref{Fig:multif}) as expected.

It is important to note that our work is a preliminary attempt to improve the capabilities of multi-spike learning rules by providing efficient alternatives. We provide systematic insight into various learning properties of our methods with synthesized experiments, while leaving possible extension to larger, more complex and practical problems unexplored in this study. This leaves a room for future developments where a proper encoding scheme is required to convert external stimuli into spikes \cite{Panzeri10,yu2013rapid,gutig2016,hopfield1995pattern,woods2018fast}.
Another limitation of our work is that only single-layer learning is examined. As layered structure is inevitable in nervous systems and has shown great importance in the success of deep learning \cite{lecun2015deep}, a potential direction for future research is thus to extend the learning capability to multi-layer structures. In spite of various early efforts \cite{Liu2019STDP,zheng2017online,cao2015spiking}, it is still valuable and important to examine the single-layered learning of spikes, in a way to closely depict its performance boundary where a complex structure would be unnecessary thanks to the computational advantages of spiking neurons \cite{gutig06,hopfield1995pattern,maass1997networks}.

\section{Conclusion}
\label{sec:Conclusion}
In this work, we proposed several new approaches towards efficient processing and learning of spikes. Firstly, we introduced a simple spiking neuron model and then converted it to an efficient form for computation where an event-driven scheme was presented. We highlighted the simplicity of our model for implementations as well as its efficiency for processing. Based on our neuron model, we proposed two efficient multi-spike learning rules, namely EML and EMLC. Our results showed that both our rules are more efficient for spike processing than other baselines. In addition to efficiency, our learning rules are highly robust to strong noise of different types. In our feature detection experiment, we re-examined the unsupervised STDP and found a new phenomenon of losing selectivity due to the domination of depression. In contrast, our proposed rules demonstrated a reliable learning over a wide range of setups without specific constraints being imposed. Moreover, our efficient learning rules can not only detect the features but also discriminate them. In summary, the simplicity, efficiency, robustness, generalization and computational power of our rules could make them a preferable choice in neuromorphic computing.

%
%\section*{Appendix}
%%You should put the details that are not required in the main body into this Appendix.

%\section*{Acknowledgment}

\ifCLASSOPTIONcaptionsoff
  \newpage
\fi

\begin{IEEEbiography}[{\includegraphics[width=1in,height=1.25in,clip,keepaspectratio]{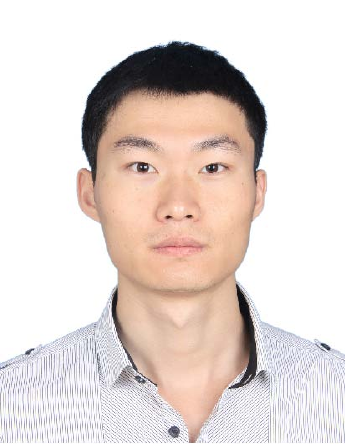}}]{Qiang~Yu}
(M'12) received the B.Eng. degree in electrical engineering and automation from the Harbin Institute of Technology, Harbin, China, in 2010, and the Ph.D. degree in electrical and computer engineering from the National University of Singapore, Singapore, in 2014.\\
He is an Associate Professor with the College of Intelligence and Computing,
Tianjin University, Tianjin, China. Before that, he was a Post-Doctoral Research Fellow with the Max-Planck-Institute for Experimental Medicine, G\"{o}ttingen, Germany, from 2014 to 2016, and a Research Scientist in the Institute for Infocomm Research, Agency for Science, Technology and Research, Singapore, from 2016. He is a recipient of the 2016 IEEE Outstanding TNNLS Paper Award. His current research interests include learning algorithms in spiking neural networks, neural coding, cognitive computations and machine learning.
\end{IEEEbiography}

\begin{IEEEbiography}[{\includegraphics[width=1in,height=1.25in,clip,keepaspectratio]{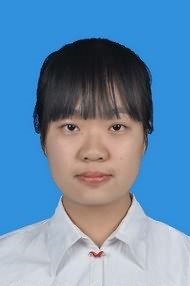}}]{Shenglan~Li}
	received her B.Sc. degree in Computer Science and Technology from Beijing University of Chemical Technology. She is currently a graduate student now studying in Tianjin University, Tianjin, China. Her current research interests include learning algorithms in spiking neural network, neural encoding and machine learning.
\end{IEEEbiography}

\begin{IEEEbiography}[{\includegraphics[width=1in,height=1.25in,clip,keepaspectratio]{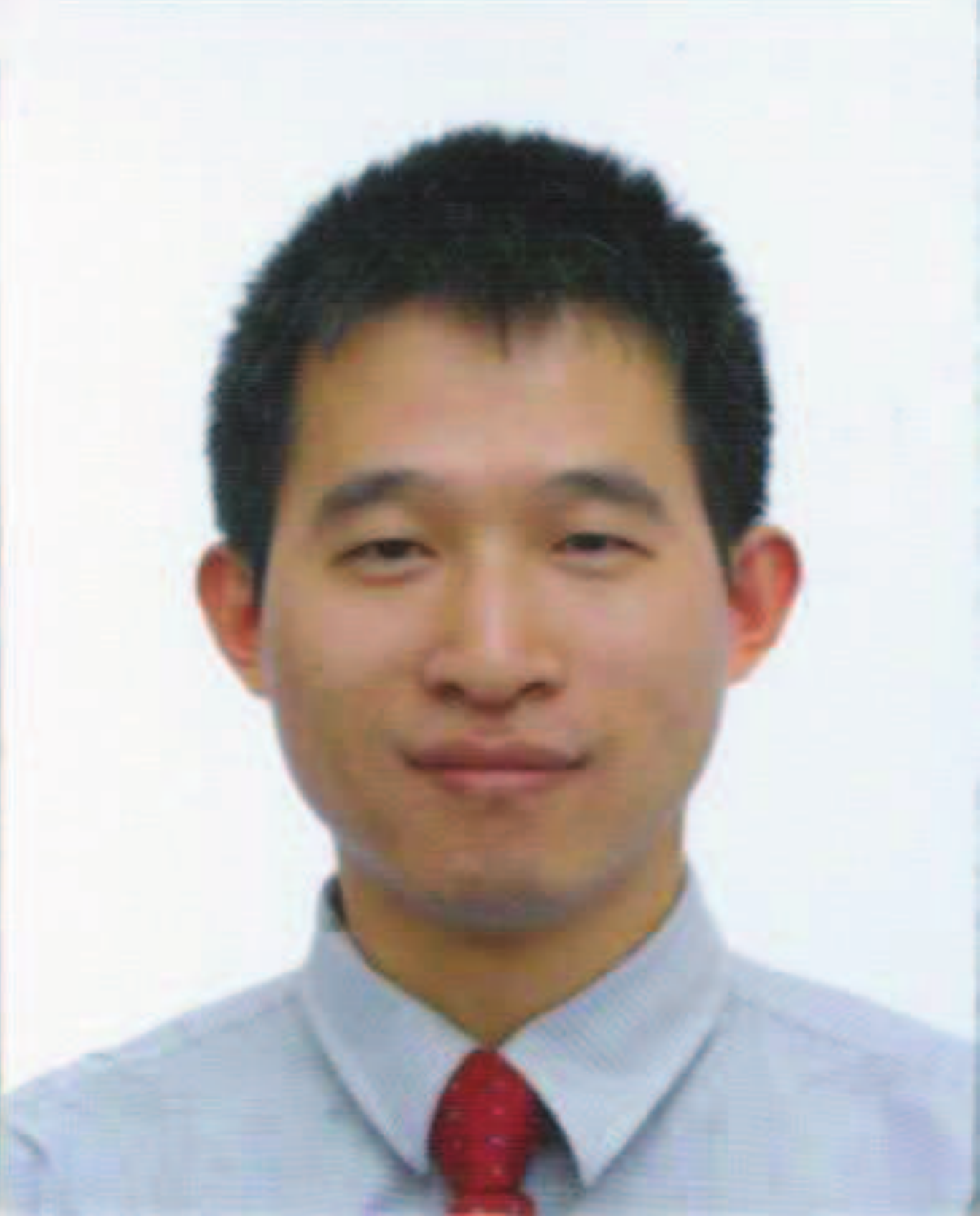}}]{Huajin~Tang}
	received the B.Eng. degree from Zhejiang University, China in 1998, received the M.Eng. degree from Shanghai Jiao Tong University, China in 2001, and received the Ph.D. degree from the National University of
Singapore, in 2005. \\
He is
currently a professor at Zhejiang University, China. His research work on
Brain GPS has been reported by MIT Technology Review
in 2015. He received the 2016 IEEE Outstanding TNNLS Paper Award. His current
research interests include neuromorphic computing, neuromorphic hardware and
cognitive systems, robotic cognition, etc. Dr. Tang has served as an Associate Editor
of IEEE Transactions on Neural Networks and Learning Systems, IEEE Transactions
on Cognitive and Developmental Systems and Frontiers in Neuromorphic Engineering. He was the Program Chair of the 6th and 7th IEEE CIS-RAM, and Chair of 2016
and 2017 IEEE Symposium on Neuromorphic Cognitive Computing.
\end{IEEEbiography}

\begin{IEEEbiography}[{\includegraphics[width=1in,height=1.25in,clip,keepaspectratio]{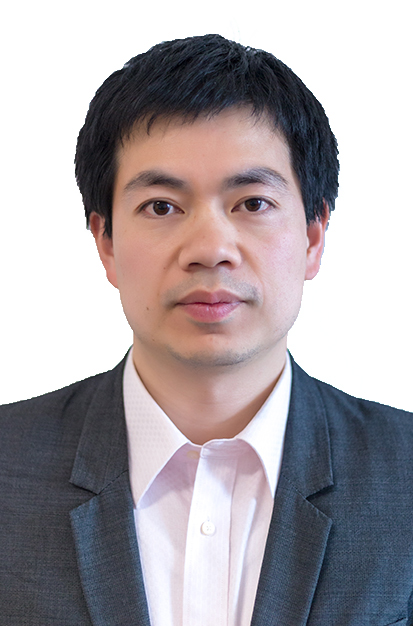}}]{Longbiao~Wang}
	received his Dr. Eng. degree from Toyohashi University of Technology, Japan, in 2008. He was
	an assistant professor in the faculty of engineering at Shizuoka University, Japan from April 2008 to September
	2012. From October 2012 to August 2016 he was an associate professor at Nagaoka University of Technology,
	Japan. Currently he is a Professor at the Tianjin University, China. His research interests include robust speech
	recognition and speaker recognition.
\end{IEEEbiography}

\begin{IEEEbiography}[{\includegraphics[width=1in,height=1.25in,clip,keepaspectratio]{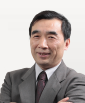}}]{Jianwu~Dang}
	(M'12) graduated from Tsinghua Univ., China, in 1982, and got his M.S. degree at the same university in 1984.  He worked for Tianjin Univ. as a lecture from 1984 to 1988. He was awarded the PhD degree from Shizuoka Univ., Japan in 1992. He worked for ATR Human Information Processing Labs., Japan, as a senior researcher from 1992 to 2001. He joined the University of Waterloo, Canada, as a visiting scholar for one year from 1998. Since 2001, he has worked for Japan Advanced Institute of Science and Technology (JAIST) as a professor. He joined the Institute of Communication Parlee (ICP), Center of National Research Scientific, France, as a research scientist the first class from 2002 to 2003. Since 2009, he has joined Tianjin University, Tianjin, China. His research interests are in all the fields of speech science including brain science, and speech signal processing.  He built MRI-based bio-physiological models for speech and swallowing, and endeavors to apply these models on clinics.
\end{IEEEbiography}

\begin{IEEEbiography}[{\includegraphics[width=1in,height=1.25in,clip,keepaspectratio]{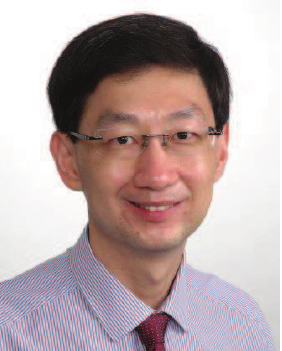}}]{Kay~Chen~TAN}
	(SM'08-F'14) received the B.Eng. (First Class Hons.) degree in electronics and	electrical engineering and the Ph.D. degree from the University of Glasgow, Glasgow, U.K., in 1994 and 1997, respectively.\\
	He is a Full Professor with the Department of Computer Science, City University of Hong Kong, Hong  Kong. He has published over 200 refereed
	articles and five books.\\
	Dr.	Tan	is the Editor-in-Chief of the IEEE TRANSACTIONS	ON EVOLUTIONARY
	COMPUTATION, was the Editor-in-Chief of the	IEEE Computational Intelligence Magazine from 2010 to 2013, and currently serves as the Editorial Board Member of over 20 journals. He is an elected member of
	the IEEE CIS AdCom from 2017 to 2019 and is an IEEE CIS Distinguished
	Lecturer from 2015 to 2017.
\end{IEEEbiography}

\end{document}